\documentclass[letterpaper, 10 pt, conference]{ieeeconf}  

\IEEEoverridecommandlockouts                              

\usepackage{graphics} 
\usepackage{epsfig} 
\usepackage{url}
\usepackage{makecell}
\usepackage{gensymb}

\title{\LARGE \bf
Multi-Platform Teach-and-Repeat Navigation by Visual Place Recognition Based on Deep-Learned Local Features*
}

\author{V\'aclav Truhla\v{r}\'ik$^{1, 2}$, Tom\'a\v{s} Pivo\v{n}ka$^{1}$, Michal Kasarda$^{1}$ and Libor P\v{r}eu\v{c}il$^{1}$  ~\IEEEmembership{Member,~IEEE} 
\thanks{This work has been submitted to the IEEE for possible publication. Copyright may be transferred without notice, after which this version may no longer be accessible. *This work was co-funded by the European Union under the project Robotics and advanced industrial production (reg. no. CZ.02.01.01/00/22\_008/0004590).}
\thanks{$^{1}$All authors are with Czech Institute of Informatics, Robotics and Cybernetics, Czech technical University in Prague, Jugoslávských partyzánů 1580/3, 160 00 Praha 6, Czech Republic ({\tt\small tomas.pivonka@cvut.cz})}
\thanks{$^{2}$V\'aclav Truhla\v{r}\'ik is also with the Department of Cybernetics, Faculty of Electrical Engineering,
Czech Technical University in Prague, Karlovo náměstí 13, 121 35 Praha 2, Czech Republic}%
}

\begin{document}
\bstctlcite{IEEEexample:BSTcontrol}

\maketitle
\thispagestyle{empty}
\pagestyle{empty}

\begin{abstract}
Uniform and variable environments still remain a challenge for stable visual localization and mapping in mobile robot navigation. One of the possible approaches suitable for such environments is appearance-based teach-and-repeat navigation, relying on simplified localization and reactive robot motion control - all without a need for standard mapping. This work brings an innovative solution to such a system based on visual place recognition techniques. Here, the major contributions stand in the employment of a new visual place recognition technique, a novel horizontal shift computation approach, and a multi-platform system design for applications across various types of mobile robots. Secondly, a new public dataset for experimental testing of appearance-based navigation methods is introduced. Moreover, the work also provides real-world experimental testing and performance comparison of the introduced navigation system against other state-of-the-art methods. The results confirm that the new system outperforms existing methods in several testing scenarios, is capable of operation indoors and outdoors, and exhibits robustness to day and night scene variations.
\end{abstract}

\section{Introduction}
\label{introduction}

Teach-and-repeat (T\&R) systems serve for the autonomous navigation of a mobile robot along a previously taught trajectory. During the teaching phase, the robot is manually (or by another system) guided along the desired trajectory while processing sensor data and recording the results. In the subsequent navigation phase, the T\&R system is able to follow the taught trajectory autonomously using the previously recorded information.

There are two core approaches to visual T\&R navigation: 
\begin{itemize}
  \item Position-based systems estimate the real robot position within the environment, and control commands are computed based on the difference between stored and measured positions. Therefore, these systems usually create a map of the environment used for precise robot localization.
  \item An alternative appearance-based approach takes advantage of fixing the  trajectory, so only 1D localization along a path or position tracking are sufficient for navigation. Instead of mapping, these systems store information from individual images during the teaching traversal. The control strategy is reactive, with heading corrections computed directly from shifts detected between stored and current images.  In many cases, the navigation is primarily based on wheel odometry, with a vision part only correcting the accumulated error.
\end{itemize}

This article focuses on the latter type of T\&R systems and introduces an innovative version of the monocular appearance-based system using visual place recognition (VPR) techniques \cite{SSM-Nav}. The new system keeps the original structure; however, it replaces both essential components of the method (VPR and computation of displacement between images) by novel approaches.  The system adopts a new VPR system \cite{SSM-VPR2} based on standard deep-learned local visual features that are further used for shift computation. Similarly to the original T\&R system, the new one inherits high robustness to various changes in the environment from the VPR method, which is a key characteristic for long-term autonomy. The system was tested indoors and outdoors and directly compared to other state-of-the-art appearance-based methods. Besides, an inseparable part of this work is design and gathering of a new public dataset made specifically for testing appearance-based techniques.

The key contributions of the new system are:
\begin{itemize}
  \item Localization along the  previously taught trajectory using the new visual place recognition method SSM-hist \cite{SSM-VPR2}
  \item Horizontal shift computation based on D2-Net local visual features \cite{D2-Net} and histogram of shifts approach used in the VPR system
  \item Targeted system design for multi-platform applications with various types of vehicles, including unmanned aerial vehicles (UAVs) that mainly do not provide precise odometry information
\end{itemize}

Despite the constraints of the T\&R scenario, these systems have several potential applications. They are suitable for various industrial tasks requiring the transportation of products or materials between several pre-defined positions. Another typical task is periodic patrolling. In addition, these systems can be used in multi-robot systems, where the robots can follow the same path, sharing the common taught trajectory.

The rest of the article is structured as follows. Sect. \ref{RW} introduces the related T\&R systems. The new T\&R system and its individual components are described in Sect. \ref{SystemDesc}. Sect. \ref{Dataset} presents a new dataset for testing appearance-based methods. The results of an experimental evaluation on the dataset and real robot are presented in Sect. \ref{experiments}. Finally, the achieved results are summarized in Sect. \ref{conclusions}.

\begin{figure}[ht]
    \centering
    \vspace{2mm}
    \includegraphics[width=0.49\linewidth]{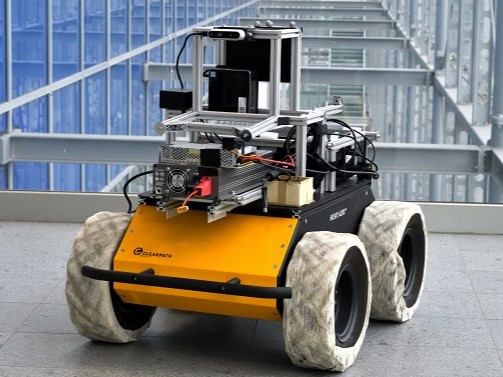}
    \hfill
    \includegraphics[width=0.49\linewidth]{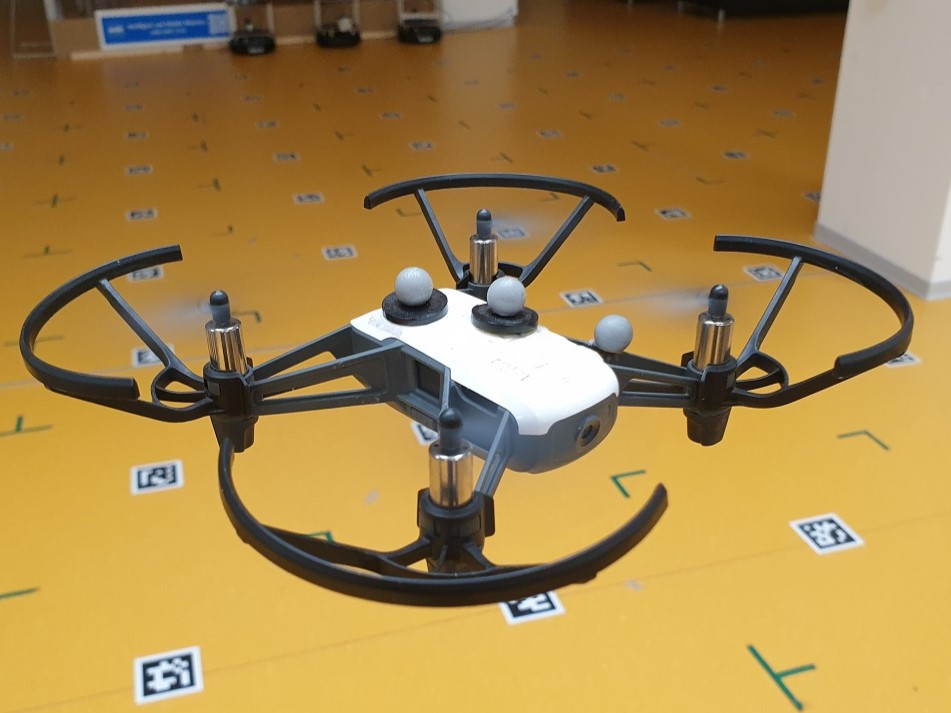}
    \caption{Husky A200 mobile robot and DJI Ryze Tello drone used for experimentation}
    \vspace{-2mm}
    \label{fig:HW}
\end{figure}

\section{Related Work}
\label{RW}

\subsection{SSM-Nav}
The T\&R system \cite{SSM-Nav}, referred to as SSM-Nav, was the first appearance-based solution using advanced VPR based on deep-learning methods for localization along the taught trajectory. Namely, the Semantic and Spatial Matching VPR (SSM-VPR) \cite{SSM-VPR} is used. This T\&R system was also one of the first systems using deep-learned local visual features.

Local visual features are directly extracted at fixed grid positions as subtensors of the inner layers of the convolutional neural network (CNN). Their size is further reduced by applying a pre-trained principal component analysis model. As the two-stage VPR is used, the features for both stages are extracted from different layers, providing features with different amounts of semantic and spatial information. Local features for the second re-ranking stage also serve to compute the shift for the reactive control.

Visual localization based on VPR was complemented by a particle filter, merging it with a driven distance measured by wheel odometry. Besides, several techniques were introduced to improve the navigation performance:

\begin{itemize}
  \item Varying distance between reference images for straight and curved trajectory segments (sampling density modulation by trajectory curvature) during teaching
  \item Adjusting forward speed for straight and curved segments during navigation
  \item Skipping the first stage of VPR, if the robot is localized with high certainty
  \item Recovery mode after localization certainty drop
\end{itemize}

This T\&R system was deployed on a four-wheel mobile robot with a forward-looking camera and tested both indoors and outdoors. It benefited from the robustness of the VPR and excelled under challenging lighting conditions. Besides, its precision was sufficient for precise indoor navigation, including passing doors.

SSM-VPR was also directly used for localization in an experimental T\&R system for a small UAV presented in \cite{SSM-drone}.

\subsection{Other Appearance-Based Systems}
The first appearance-based systems were introduced in the 1990s \cite{VSRR}, followed by methods based on local visual features in the next decade \cite{qual_vis_path_follow_orig}, \cite{Krajnik2010}. These two systems were designed only for polygonal trajectories. A more advanced solution based on local features and working with general trajectories is the BearNav system \cite{BearNav}.

A novel CNN approach to compute the horizontal shift between images was introduced in \cite{Rozsypalek-22}. Both images are processed by the same CNN (Siamese network), returning 3D tensors. One of the tensors is extended by padding in the direction corresponding to the horizontal shift, followed by the convolution of these tensors. The output is a vector with the likelihoods of possible displacements, and the displacement with the highest likelihood is the final shift. This architecture was further used in \cite{Roucek-22} to support local feature matching in the BearNav system, and applied in the system with multi-dimensional particle filter \cite{Rozsypalek-23}.

The CNN-based method using Siamese network \cite{Rozsypalek-22} along with the original Bearnav approach \cite{BearNav}, both implemented in the new Bearnav2 system \cite{BearNav2}, were used as reference systems for the experimental evaluation in this work.

\subsection{Position-Based Systems}
The fundamental work for the position-based T\&R navigation is \cite{Barfoot2010} from 2010, presenting a system capable of following several kilometers long trajectories. The system is based on SURF local features \cite{SURF} and uses a stereo camera to build a sequence of local maps. This approach was further modified for a monocular camera \cite{Barfoot2016_mono}, multi-experience localization \cite{Barfoot2016_appearance, Barfoot2017_triage}, and deep-learned local features \cite{TaR-features} detected by a U-Net architecture.

T\&R system directly based on a standard simultaneous localization and mapping system was presented in \cite{ORB2TaR}. Several position-based systems have also been introduced for UAVs \cite{Barfoot-drone, Nitsche_vio}.

\section{Teach-and-Repeat Navigation System}
\label{SystemDesc}

\begin{figure}[b]
    \centering
    \includegraphics[width=\columnwidth]{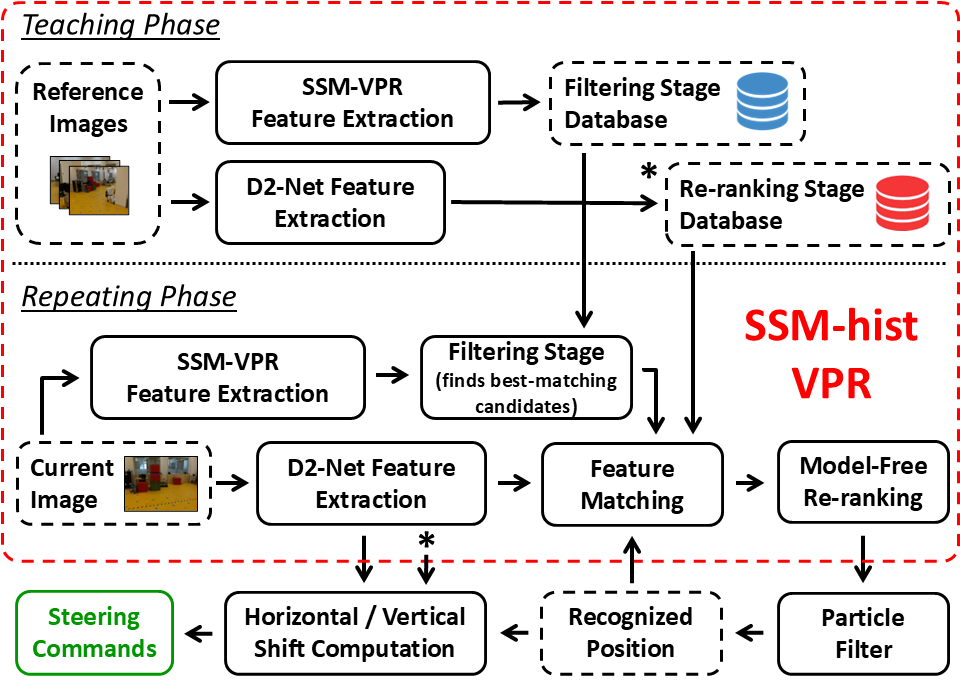}
    \caption {Structure of the introduced T\&R system SSM-Nav2}
    \label{fig:flowchart}
\end{figure}

The introduced T\&R system, adopting design from \cite{SSM-Nav}, is divided into two separate parts for teaching and navigation. The flowchart presented in Fig. \ref{fig:flowchart} depicts the main components of both parts, and the following subsections describe individual steps in more detail.

\subsection{Teaching Phase}
During the teaching phase, the system continuously captures images, extracts local visual features, and stores them in the database. For each image, it also stores the driven distance measured by wheel odometry. If the odometry is unavailable, the distance is estimated from the operator's steering commands. The VPR system uses two-stage recognition with filtering, quickly preselecting the best candidates from the whole database, and subsequent re-ranking, performing more detailed comparisons. The system extracts local features for both stages separately; specifically one type of SSM-VPR \cite{SSM-VPR} features for the filtering and D2-Net local visual features \cite{D2-Net} for the re-ranking. D2-Net features are also used for the computation of the shift between images.

Another essential component of the teaching part is the image-capturing strategy. The images should properly represent the whole trajectory, but the high number of images negatively influences the localization speed, especially during initialization, and increases memory requirements. As stated in \cite{SSM-Nav}, an adaptive strategy is more convenient than the constant trigger distance. The system decreases the distance while the robot turns and increases it on straight parts of the trajectory with less significant changes between images. This distance is measured and controlled based on wheel odometry (or estimated from steering commands).

Optionally, the system computes and stores horizontal shifts between subsequent reference images, as described in Sect. \ref{sect:shift-comp}. These shifts can improve the navigation, providing additional information to the reactive control that detects the current shift only.

\subsection{Visual Place Recognition}
\label{SystemDesc-VPR}

For the localization along the trajectory, the presented T\&R system uses the new VPR system introduced in \cite{SSM-VPR2}. This system adopts the filtering stage from the SSM-VPR \cite{SSM-VPR} and employs new re-ranking techniques using correspondences of D2-Net local visual features \cite{D2-Net}. For the application in the T\&R system, the histogram of shifts technique \cite{SSM-VPR2}, achieving the best results from the introduced methods, was selected.

For each image pair, the histogram of shifts method creates an empty two-dimensional (2D) histogram of all potential shifts between matched local features. The shift is a 2D vector computed directly as a difference of image coordinates of matched features. These matches are found as the mutually nearest neighboring features (cross-check). Subsequently, all feature matches vote in this histogram by their shifts. They do not vote only for their particular bins but for all bins using Gaussian weighting, which helps to reduce inconsistencies caused by imprecise detection and projection geometry. In addition, each vote is weighted by the match score. Here, the method benefits from the used D2-Net local feature detector \cite{D2-Net} returning a score for each feature, so the match weight is computed as a sum of both features' scores. After all matches have voted, the final similarity score is found as a maximum value from the histogram.

In the re-ranking stage, similarity scores of the current query image to all candidate images found in the filtering stage are computed. Based on these scores, the candidates are re-ranked, and the best matching image representing the recognized position is retrieved. In the presented T\&R system, this position is not directly used for the computation of the steering command, but is further filtered, as described in the next subsection, to suppress incorrect recognitions.

\subsection{Position Filtering}

The currently localized position is tracked by a particle filter, joining visual localization and odometry, similarly to \cite{SSM-Nav}. At the beginning, particles are uniformly initialized along the whole trajectory. The motion model regularly shifts the particles by the distance traveled since the last update with a small added noise. At each step, the filter discards the particles out of the trajectory or with the lowest weight and initializes new ones.

When new sensor information from VPR is received, neighboring database image positions are assigned to each particle. Subsequently, the weight of each particle is updated with the weights of its two neighboring images, interpolated based on the distance to them. Image weights are the scores from the re-ranking stage of the VPR if the image belongs among the best-found candidates. Otherwise, the score is zero. The final position is retrieved as an average of the five best particles' positions.

The weights of particles are further used to switch between standard navigation and initialization mode, which is used when the robot is localized with high uncertainty. Contrarily, when the localization certainty is high, the system can temporarily turn off the first stage of the VPR.

\subsection{Horizontal Shift Computation}
\label{sect:shift-comp}

Horizontal shift between two images is computed using the histogram of shift technique from VPR, presented in Sect. \ref{SystemDesc-VPR}. It is retrieved almost equally to the similarity score, but instead of the maximum value, the system returns the horizontal shift for which this maximum value was reached. As the histogram is two-dimensional, the system simultaneously computes also the vertical shift needed for the navigation of aerial vehicles.

In the presented T\&R system, the horizontal shift is computed for the current image and the image from the database, which is closest to the localized position along the trajectory. If the precomputed shifts between database images are used, the measured shift is directly averaged with the stored shift. Then, the final shift is transformed into the angular velocity command using a basic proportional regulator. The only additional operation is a transformation of the horizontal shift into the angular based on the known focal length of the camera.

\subsection{System for Aerial Vehicles}
\label{sect:aerial}

The presented T\&R system is capable of navigating UAVs using the following modifications. Wheel odometry is replaced by an estimate of a pose using known velocity commands. Despite the lower precision, this approach sufficed for most of the testing scenarios, as the odometry is used only for measuring the relative traveled distance, and the localization primarily relies on visual information.

The second modification is the altitude control based on the vertical shift between images. As mentioned in the previous section, the vertical shift is returned simultaneously with the horizontal one. Therefore, the system was only extended with an additional proportional regulator, transforming the vertical shift into a linear velocity.

\section{Dataset for Testing Appearance-Based Teach-and-Repeat Methods}
\label{Dataset}

\subsection{Dataset Description}
T\&R systems are usually tested for traversing particular trajectories in real experiments, evaluating their overall performance. The used metrics measure navigation precision or the capability to traverse the whole trajectory. However, the overall performance depends on many factors, including the tuning of various parameters. Therefore, testing particular approaches separately on standardized datasets can be more convenient for their exact comparison. For the presented system, these approaches are especially VPR-based localization and horizontal shift computation. The standalone VPR is not tested in this article since the evaluation of the used VPR methods on many public datasets was already published in \cite{SSM-VPR2}. In addition, imprecision of the recognition can be suppressed by position filtering supported by odometry.

By contrast, to the best of our knowledge, there is no public dataset specifically designed for horizontal shift estimation, yet. Besides, the applicability of other existing datasets designed for VPR or other mobile robotic tasks (e.g., visual odometry) is limited in this case. These datasets often miss multiple views from the same place, mutual camera positions are unknown, or they do not represent the right transformations (i.e., lateral shift and horizontal rotation). Therefore, we decided to create a new unique dataset suited directly for this task.

The new dataset contains three image sequences comprising sets of images taken from subsequent positions on the tested trajectories. The set always provides 9 images with mutually known transformations for each position. These are combinations of three lateral shifts with three horizontal camera rotations for each capturing point. The central forward-looking positions simulate teaching phase images, and the other images can simulate the repeating phase. The dataset was captured manually, moving the camera on a metal construction that ensured millimeter precision of the lateral shift and rotation error below 2 \degree.

\begin{figure}[hb]
    \centering
    \includegraphics[width=0.32\linewidth]{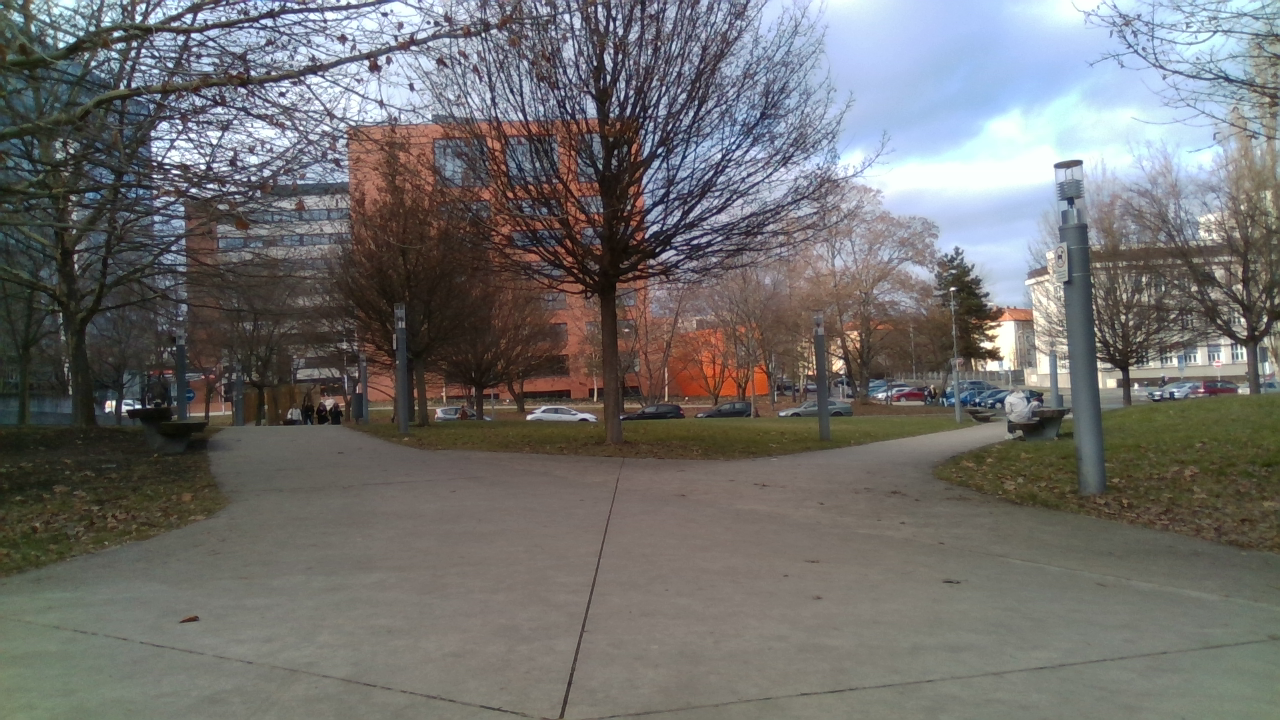}
    \hfill
    \includegraphics[width=0.32\linewidth]{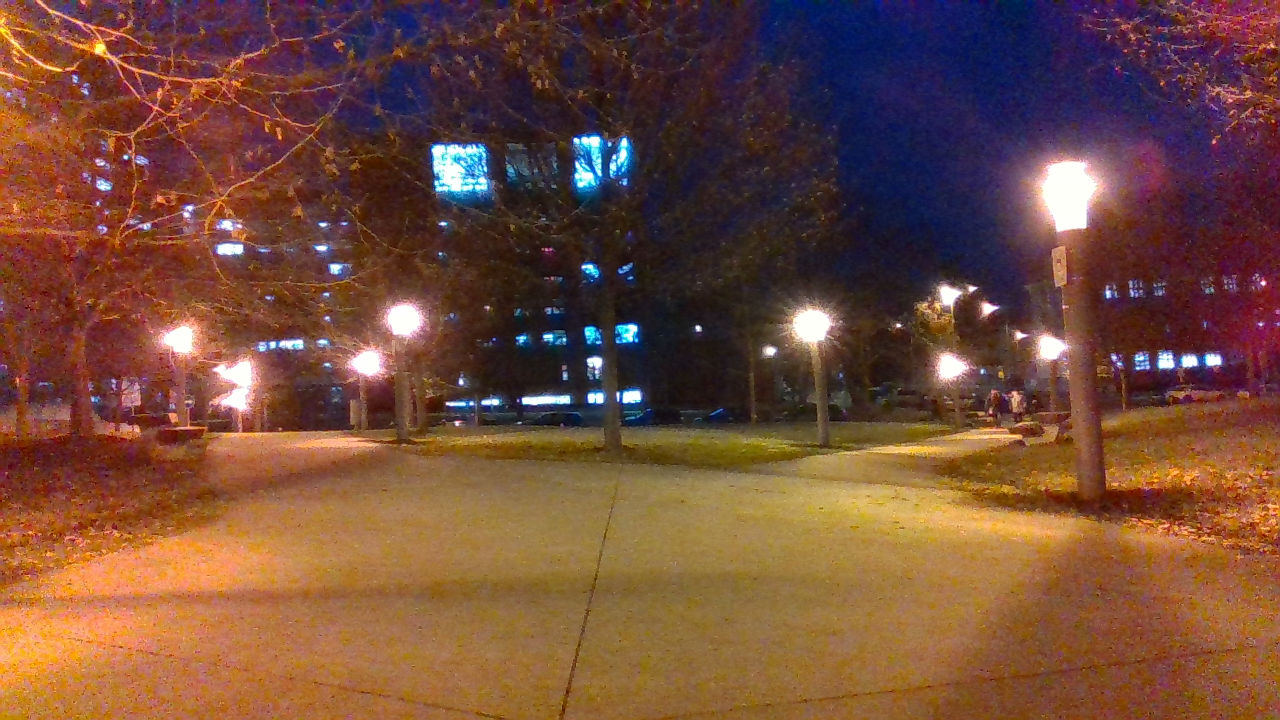}
    \hfill
    \includegraphics[width=0.32\linewidth]{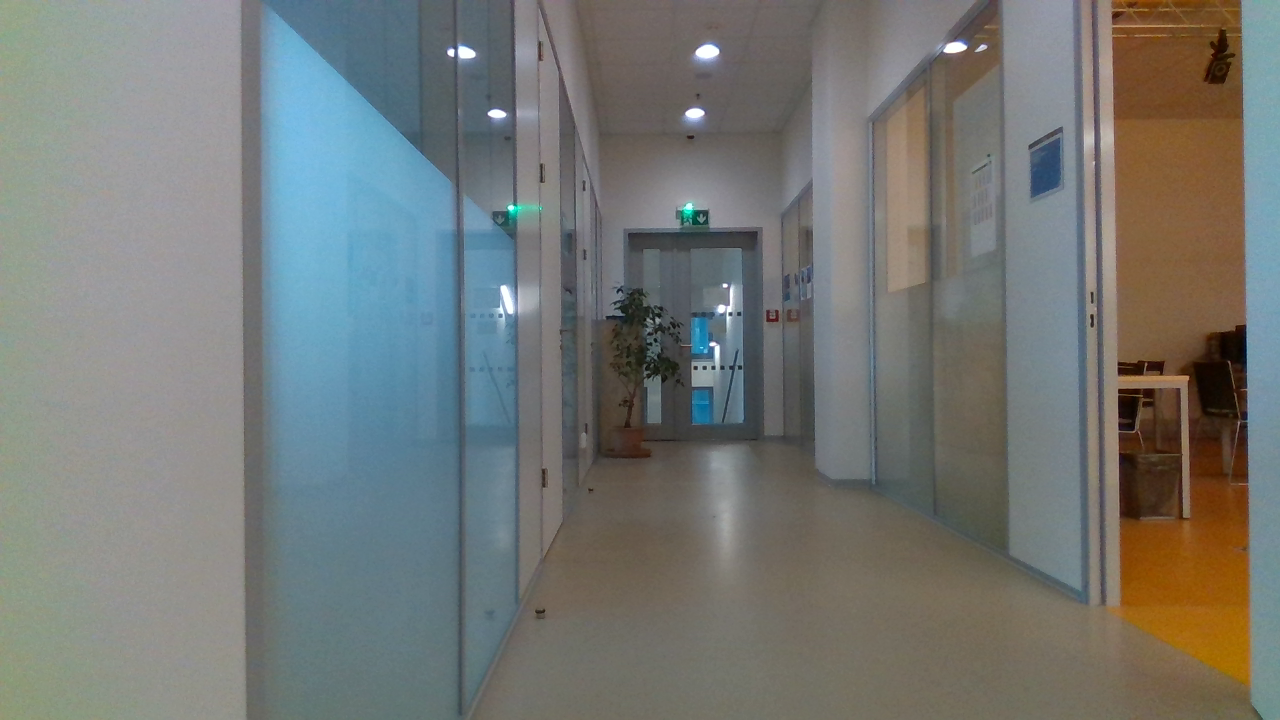}
    \vspace{-1mm}
    \caption{Example images of new dataset for shift computation testing}
    \label{fig:Dataset-imgs}
    \vspace{-4mm}
\end{figure}

Two image sequences were taken outdoors on the same trajectory on a university campus, mixing urban and natural environments (Fig. \ref{fig:Dataset-imgs}). The first sequence was recorded during the day, followed by the second one a few hours later in the evening after sunset to reflect different lighting conditions. The images were taken always from identical positions for both sequences, so they can be combined. The third sequence was captured indoors in the corridors of a university building, containing repetitive structures and many uniform and low-textured surfaces.

Each outdoor sequence contains 459 images from 51 positions, and the indoor sequence has 279 images from 31 positions. The lateral shifts from the central forward-looking position were 36 cm in both left and right directions. The horizontal camera rotations were $\pm 15 \degree$. The camera used was an Intel RealSense D435 with a resolution of 1280 x 720 px.

The dataset is publicly available at \url{https://imr.ciirc.cvut.cz/Datasets/TaR}.

\subsection{Metrics}
\label{sec:Metrics}
Appearance-based systems usually return a single shift between two images in image coordinates. Although the real transformation between the images is known, it is not possible to determine a single correct ground-truth value. The correct value varies for projections of different 3D points, so the singe returned value tends to simplify in reality more complex projective geometry. Therefore, the following metrics for comparing the methods were designed.

For the first metric, the correct shifts of two 3D points lying on the reference camera axis are computed. The first point was set at a distance of 5m from the camera center, and the second lies at the infinity. The metric evaluates how many shifts (database images) for a particular transformation fit into the range between shifts of these two reference points. This range was slightly enlarged to add a tolerance for almost correct shifts (20 px). This is especially important in cases where the camera rotates around its center so the shifts for both reference points are equal.

The second additional metric evaluates the consistency of the returned shift for equal transformations, expressed directly by the standard deviation.

For the presented dataset, these metrics were computed separately for individual transformations to the central forward-looking position and then averaged to evaluate the overall performance.

\section{Experiments}
\label{experiments}

\subsection{Testing of Horizontal Shift Computation}
\label{sect:shift_test}
At first, T\&R systems were tested using the collected dataset presented in Sect. \ref{Dataset}. The new SSM-Nav2 system based on D2-Net local visual features \cite{D2-Net} was compared with its earlier version SSM-Nav \cite{SSM-Nav} and two variations of the BearNav2 system \cite{BearNav2} as a reference, using the Siamese network and SIFT local features. The systems were tested with all 3 dataset trajectories captured outdoors during the day, outdoors at night, and indoors. Since day-time and night datasets can be combined, day-time images were used as a reference for night images. For the performance evaluation, the metrics from Sect. \ref{sec:Metrics} were used.

The results presented in Tab. \ref{Tab:dataset-results} show significant superiority of the presented system, achieving the best performance under all conditions. All deep-learning-based solutions also outperform the traditional vision method using SIFT features under night conditions.

\begin{table}[t]
    \centering
    \vspace{2mm}
    \caption{Shift computation testing on day-time, night and indoor sequences of a new dataset using metrics from Sect. \ref{sec:Metrics}}
    \renewcommand{\arraystretch}{2.2}
    \begin{tabular}{|l|c|c|c|c|c|c|}
    \hline
     & \makecell{ Correct \\ Shift \\ Day \\{[\%]}} & \makecell{ STD \\ Day \\{[px]}} & \makecell{ Correct \\ Shift \\ Night \\{[\%]}} &  \makecell{ STD \\ Night \\{[px]}} & \makecell{ Correct \\ Shift \\ Indoor \\{[\%]}} &  \makecell{ STD \\ Indoor \\{[px]}} \\ \hline
    \makecell{BearNav2 \\ SIFT} 
    &  99 & 7 & 33 & 233 & 81 & 64 \\ \hline
    \makecell{BearNav2 \\ Siam}
    &  99 & \textbf{6} & 90 & \textbf{17} & 79 &  33\\ \hline
    SSM-Nav
    &  88 & 9 & 74 & 80 & 68 & 37 \\ \hline
    \makecell{\textbf{SSM-Nav2} \\ \textbf{D2-Net}}
    & \textbf{100} & 10 & \textbf{96} & 19 & \textbf{83} & \textbf{27} \\ \hline
    \end{tabular}
  \label{Tab:dataset-results}
  \vspace{-4mm}
\end{table}

\subsection{Hardware and System Setup}

During experiments, all T\&R systems were tested on a four-wheel differential drive robotic platform Clearpath Husky A200. This mobile robot was additionally equipped with an Intel RealSense D435 camera, offering a maximum resolution of 1920 x 1080 px, and an external computer. The computer comprised an Nvidia GeForce GTX 1070 graphics card, an Intel Core i5-7300HQ processor, 32 GB of RAM, and a 500 GB SSD. This setting was identical to the experiments presented in \cite{SSM-Nav}.

The software part of the presented T\&R system was built on ROS2 Humble and Ubuntu 20.04. Individual ROS nodes were developed in Python. For the experiments, the original SSM-Nav system was also re-implemented for this setting, additionally using PyTorch library instead of Keras similarly to the newer versions of the SSM-VPR system \cite{SSM-VPR}.

The used image resolutions in pixels for particular T\&R systems were 336 x 336 for the SSM-Nav2, 448 x 448 for the original SSM-Nav (both with 224 x 224 for the VPR filtering stage) and 1280 x 720 for BearNav.

\subsection{Indoor Experiments}

\begin{figure}[b]
    \centering
    \vspace{-3mm}
    \includegraphics[width=\columnwidth]{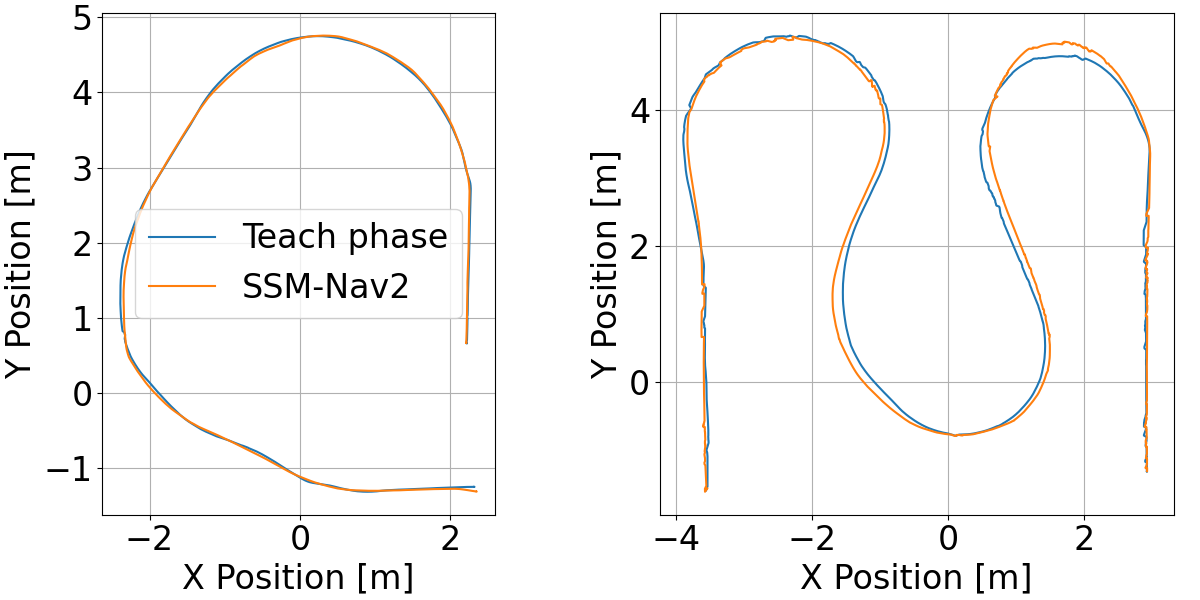}
    \caption{Basic (left) and Advanced (right) trajectories driven by SSM-Nav2}
    \label{fig:trajectories}
\end{figure}

\begin{table*}[t]
    \centering
    \vspace{2mm}
    \caption{Navigation precision of T\&R systems for indoor experiments on two testing trajectories, 
    including various lighting conditions, laterally shifted starting position (30 cm) and starting in the middle of the trajectory.
    }
    \vspace{-2mm}
    \renewcommand{\arraystretch}{1.3}
    \begin{tabular}{|l|c|c|c|c|c|c|c|c|}
    \hline
    & \multicolumn{5}{|c|}{Basic Path} & \multicolumn{3}{|c|}{Advanced path} \\ \hline
    & \multicolumn{2}{|c|}{Normal Start} & Shifted Start & Start in Middle & Night Run & \multicolumn{2}{|c|}{Normal Start} & Night Run \\ \hline
    & \makecell{ Mean \\ Deviation \\{[cm]}} & \makecell{ Maximum \\ Deviation \\{[cm]}} & \makecell{Mean \\ Deviation \\ {[cm]}} & \makecell{ Mean \\ Deviation \\{[cm]}} & \makecell{ Mean \\ Deviation \\{[cm]}} & \makecell{ Mean \\ Deviation \\{[cm]}} & \makecell{ Maximum \\ Deviation \\{[cm]}} & \makecell{ Mean \\ Deviation \\{[cm]}} \\ \hline
    BearNav2 Siam &  \textbf{4.2} & \textbf{8.7} & 19.9 & 79.0 & \textbf{4.7} & \textbf{3.0} & \textbf{9.1} & \textbf{3.5} \\ \hline   
    SSM-Nav 
    & \textbf{4.2} & 11.4 & 13.9 & 6.5 & 31.0 & 5.3 & 16.4 & 12.3 \\ \hline
    SSM-Nav2 (D2-Net) & \textbf{4.2} & 13.1 & \textbf{13.0} & \textbf{5.1} & 10.0 & 6.6 & 26.4 & 11.2 \\ \hline
   SSM-Nav2 (D2-Net) + Shift & 4.5 & 12.1 & 19.9 & 6.7 & 10.8 & 8.4 & 34.8 & 10.8 \\ \hline
    \end{tabular}
  \vspace{-5mm}
  \label{Tab:indoor_day}
\end{table*}

Indoor experiments were carried out in a robotic laboratory equipped with the Vicon localization system providing ground-truth positions of the robot during traversals. In contrast to the experiments in Sect. \ref{sect:shift_test}, the tested systems included besides SSM-Nav2 and SSM-Nav only the better version of BearNav \cite{BearNav} based-on the Siamese network. The SSM-Nav2 was tested in two versions with and without using the precomputed shift. All systems were tested on two different trajectories. The first one was a trajectory with a basic shape and a length of almost 17 m. The second one was more curved, with a length of almost 30 m.

During the first experiment, the systems independently repeated both taught trajectories exactly from the starting positions. In the second experiment, the robot started at the basic trajectory from the position 30 cm laterally shifted to the left. The final results for both experiments averaged over three passes for each system are presented in Tab. \ref{Tab:indoor_day}, and the recorded traversals for SSM-Nav2 from the first experiment are shown in Fig. \ref{fig:trajectories}.

All systems demonstrated the ability to follow the trajectory and to get back on the taught path during their first turn in the second experiment. However, they all converged very slowly to the original trajectory on straight segments, which is typical for appearance-based systems.
The mean and maximum deviations from the taught path shown in Tab. \ref{Tab:indoor_day} prove that the new system SSM-Nav2 is comparable to the previous version and other state-of-the-art systems as well. The experiments also showed that using precomputed shifts did not improve the system performance. This technique would be more important for systems without position tracking as \cite{SSM-drone} since VPR sometimes tends to prefer images with the same orientation, leading to ignoring turns. The higher precision of the BearNav system is primarily ensured by repeating the taught velocities.

The third experiment focused on systems' ability to start from an arbitrary position on the taught trajectory. In this category, both versions of the SSM-Nav system successfully localized itself and followed the rest of the path. By contrast, BearNav2 is not able to fulfill this task since it misses the localization and replays all commands from the teaching phase.

\begin{figure}[b]
    \centering
    \vspace{-3mm}
    \includegraphics[width=0.4\linewidth]{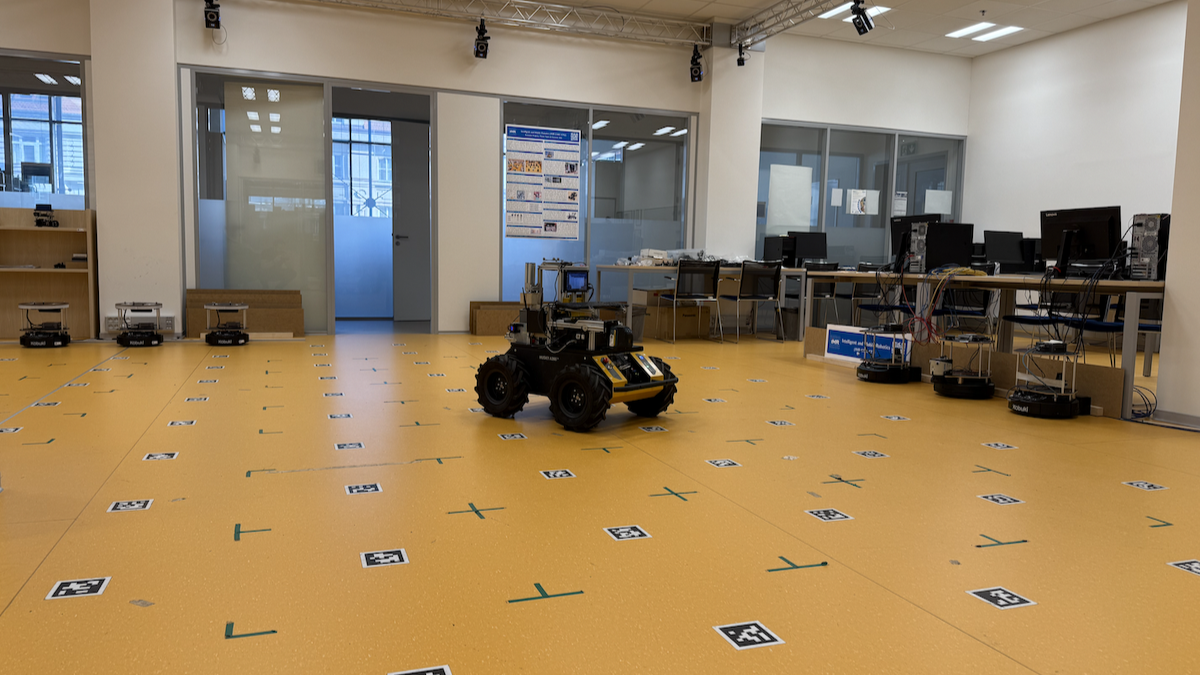}
    \hspace{1mm}
    \includegraphics[width=0.4\linewidth]{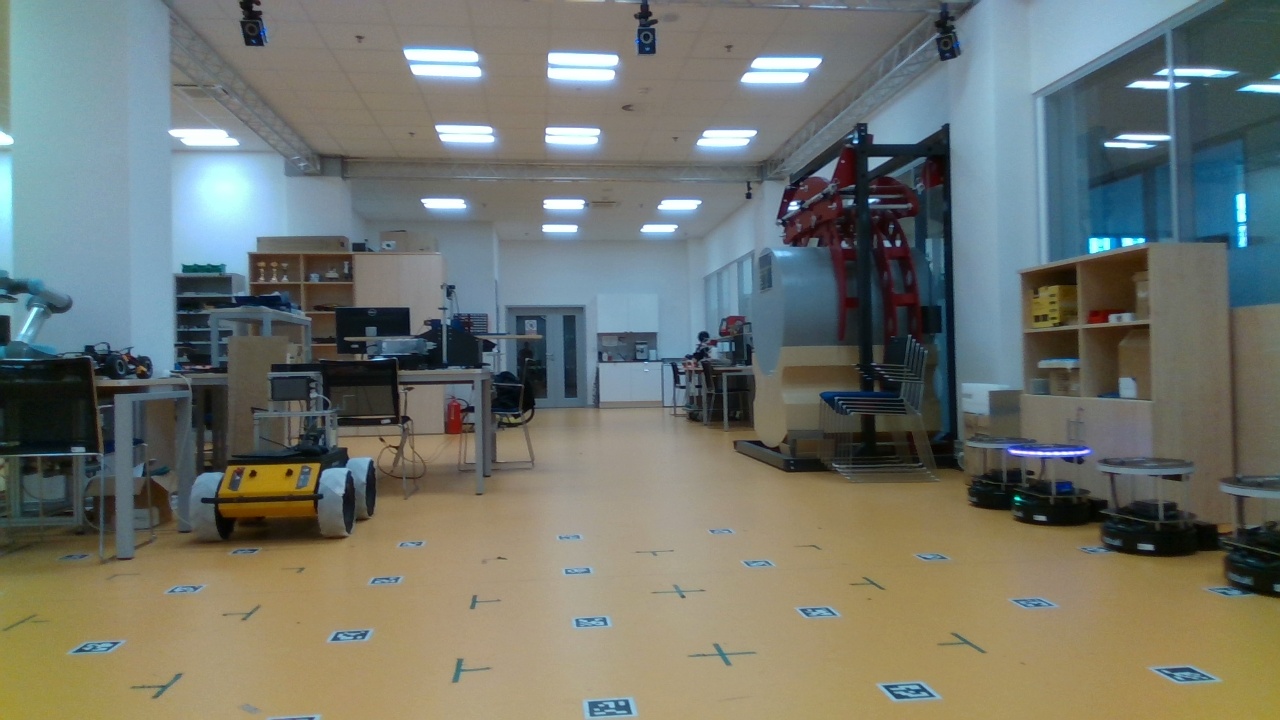}\\
    \vspace{1mm}
    \includegraphics[width=0.4\linewidth]{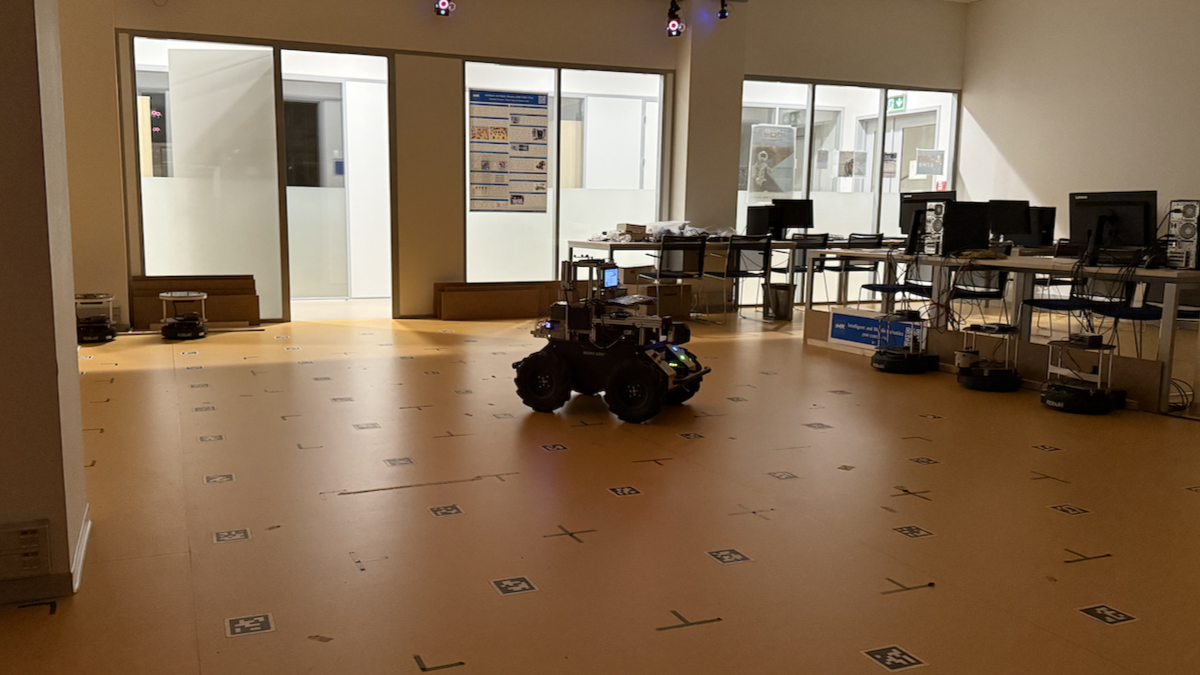}
    \hspace{1mm}
    \includegraphics[width=0.4\linewidth]{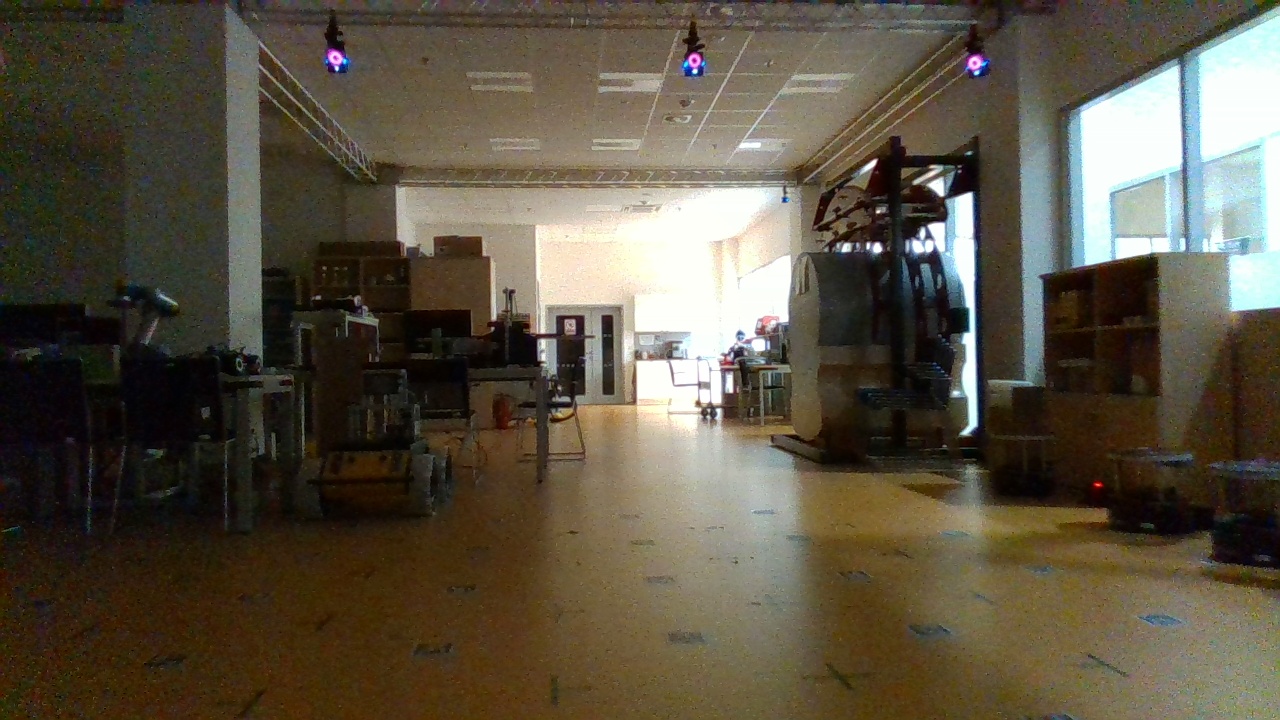}\\
    \vspace{-2mm}
    \caption{Example images from external (left) and onboard (right) cameras for day-time reference trajectory and switched-off lights during night-time}
    \label{fig:night}
\end{figure}

The final experiment was conducted on both trajectories at night time with switched-off lights, so the laboratory was dark, illuminated only by lights from an outside corridor. All tested systems were able to repeat the paths even in limited lighting conditions. Example images from this experiment, captured by an external camera and an onboard camera, are in Fig. \ref{fig:night}, and the measured errors are presented in Tab. \ref{Tab:indoor_day}.

\subsection{Navigation of Unmanned Aerial Vehicles}

A similar indoor experiment as for the ground robot was repeated for the SSM-Nav2 system remotely controlling a small UAV DJI Ryze Tello. The system was able to successfully navigate the drone along the taught path in 3D space, calculating both the horizontal and the vertical shift as described in Sec. \ref{sect:aerial}.

\section{Conclusions}
\label{conclusions}
The introduced T\&R system SSM-Nav2 features many characteristics crucial for long-term autonomy, such as the ability to work under challenging and varying lighting conditions. Another innovative property is its multi-platform design with no ultimate dependency on external odometry, which makes the system applicable to various types of vehicles, including UAVs.

The system was tested against other state-of-the-art solutions with a newly designed and gathered public dataset, targeting specifically appearance-based systems. The dataset comprises sequences of day and night outdoor images and a low-structured indoor environment. The presented system achieved the best performance over the reference solutions with this data set. Besides, the system was compared to other solutions in experiments with real robots, outperforming the original SSM-Nav system in terms of precision and speed.

Future investigations on the topic will primarily focus on correcting lateral displacements to speed up convergency to the taught path. Furthermore, additional extensive experiments should be conducted outdoors and along longer trajectories. The extension of the introduced dataset with new environments, seasons, or weather conditions for this work and public use is also targeted.






\bibliographystyle{IEEEtran}
\bibliography{bibliography}

\end{document}